\documentclass[runningheads]{llncs}
\usepackage{graphicx}

\usepackage{cite}
\usepackage{amsmath,amssymb,amsfonts}
\usepackage{algorithmic}
\usepackage[linesnumbered,ruled]{algorithm2e}
\usepackage{graphicx}
\usepackage{textcomp}
\usepackage{xcolor}

\DeclareMathOperator{\E}{\mathbb{E}}

\begin{document}
\title{An Adaptive Deep Learning Algorithm Based Autoencoder for Interference Channels}
%
%
\author{Dehao Wu\inst{1} \and
Maziar Nekovee\inst{1,2} \and
Yue Wang\inst{3}}
\authorrunning{D. Wu et al.}
\titlerunning{D. Wu et al.}
%
\institute{Centre for Advanced Communicastions, Mobile Technology and IoT, University of Sussex,  Brighton, United Kingdom \and Quantrom Technologies LTD \and
Samsung Electronics R\&D Institute, Communications House, South Street, Staines, Middlesex TW18 4QE United Kingdom \\
\email{\{dehao.wu,m.nekovee\}@sussex.ac.uk; yue2.wang@samsung.com}}
\maketitle              
\begin{abstract}
Deep learning (DL) based autoencoder (AE) has been proposed recently as a promising, and potentially disruptive Physical Layer (PHY) design for beyond-5G communication systems. Compared to a traditional communication system with a multiple-block structure, the DL based AE provides a new PHY paradigm with a pure data-driven and end-to-end learning based solution. However, significant challenges are to be overcome before this approach becomes a serious contender for practical beyond-5G systems. One of such challenges is the robustness of AE under interference channels. In this paper, we first evaluate the performance and robustness of an AE in the presence of an interference channel. Our results show that AE performs well under weak and moderate interference condition, while its performance degrades substantially under strong and very strong interference condition. We further propose a novel online adaptive deep learning (ADL) algorithm to tackle the performance issue of AE under strong and very strong interference, where level of interference can be predicted in real time for the decoding process. The performance of the proposed algorithm for different interference scenarios is studied and compared to the existing system using a conventional DL-assist AE through an offline learning method. Our results demonstrate the robustness of the proposed ADL-assist AE over the entire range of interference levels, while existing AE fail to perform in the presence of strong and very strong interference. The work proposed in this paper is an important step towards enabling AE for practical 5G and beyond communication systems with dynamic and heterogeneous interference.

\keywords{Deep learning  \and physical layer \and autoencoder \and interference channel.}
\end{abstract}
\section{Introduction}
Communication networks and services are becoming more intelligent with the novel advancements and unprecedented levels of computational capacity that is available for processing locally or in the cloud. AI,  including machine learning (ML) and deep learning (DL), has been widely used for the design and management of communication systems, and has been shown to significantly enhance the system performance and reduce the operational cost, hence has raised great interest in standard [1], as well as in research. There has been a number of examples of using AI  in communication systems in the literature, for example, for channel estimation [2], complex multiple-input and multiple-output (MIMO) detection [3], channel decoding [4], joint channel estimation and detection [5], joint channel encoding and source encoding [6].

In a conventional communication system, the channel propagation is often modeled mathematically, which may not correctly reflect the channel in practical scenarios and the dynamic nature of the changing. DL based approaches demonstrate a useful and insightful way of fundamentally rethinking the communication system design problem and hold the promise for performance enhancement in complex scenarios that are difficult to characterize with tractable mathematical models. Compared to a traditional communication system with a structure consisting multiple functional blocks, autoencoder provides a new paradigm with a pure data-driven and end-to-end learning based solution. For example, a DL based AE is proposed in [7], where the deep neural networks (DNNs) based reconstruction transceiver block jointly optimizes all the functions in a single process. The work in [8] presents end-to-end learning of a communications system without a channel model. In [9], authors propose a deep reinforcement learning approach for training a link with noisy feedback, for both additive white Gaussian noise (AWGN) and Rayleigh block-fading (RBF) channels. Also, two practical DL-based systems are implemented in [10] and [11].

\begin{figure*}[t!]
\begin{center}
\includegraphics[width=.75\textwidth]{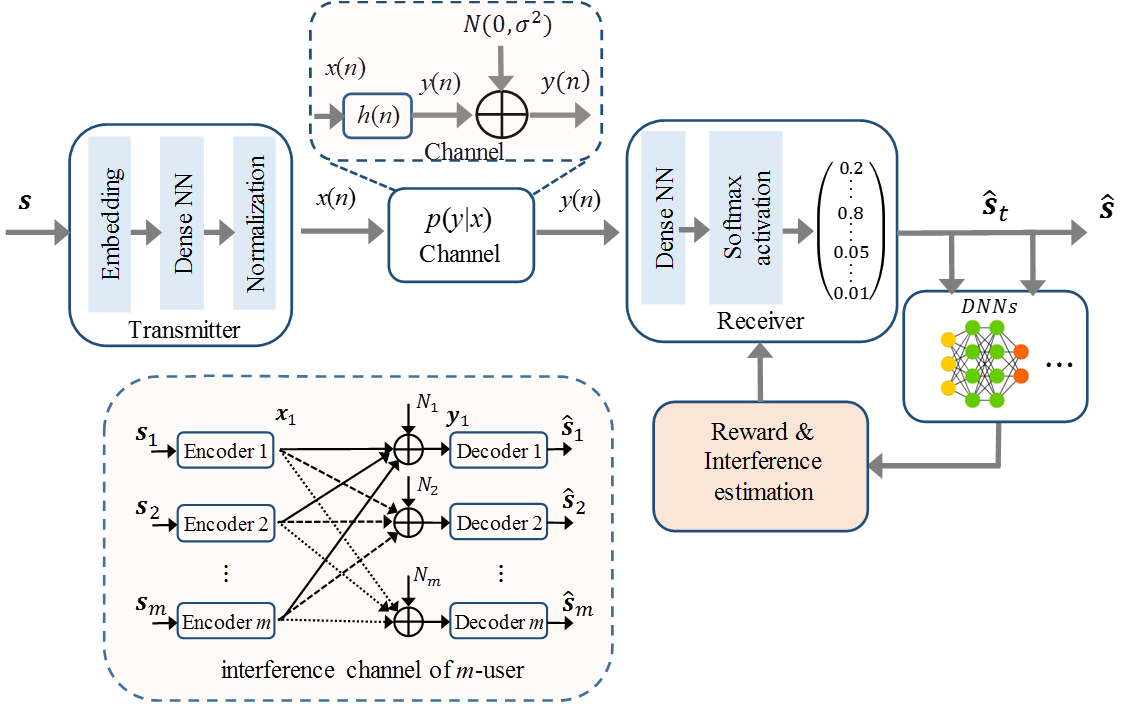}
\end{center}
\caption{System block diagram of an ADL algorithm based AE for a wireless communication interference channel with $m$-user}
\label{fig:System_block_diagram}
\end{figure*}

All the work above provided great insights of the potential performance of applying AE for interference-free channels. However, it is also revealed in [12] that AE can be vulnerable to adversarial and jamming attacks, compared to conventional coding schemes.  While [13] shows that such drawbacks can be mitigated through adversarial training, it is not clear how AE will behave under a multi-user interference channel, with which performance of a multi-user system is often impaired [14, 15]. The study in [7] considers a two-user link with interference for AE. However, offline training is used and there is no adaptive training for different levels of interference. Other studies on AE, MIMO channel learning [16], channel estimation in an OFDM system [17], and learning to optimize for interference management [18], are all based on offline learning, therefore does not cope well with the situation when interference is dynamic, can be from different sources, and can vary in real time. 

In this work, we characterize the tolerance of a conventional AE under a Gaussian interference channel, with respective to different interference levels. Our results demonstrate that although the offline trained AE approach has reasonable robustness for noisy to moderate interference channel, performance of AE suffers substantially under a strong or very strong interference channel. To date, there has been little work on DL-based AE in the presence of an interference channel with a variety of interference strengths, even less so to address the issue for allowing AE in practical dynamic and heterogamous interference scenarios. In this paper, we proposed an adaptive deep learning (ADL) algorithm based AE. The interference strength is predicted through an adaptive deep learning process, where real time online learning is performed to obtain the knowledge of the real time interference level for the subsequent decoding process, through an updated DNNs layer. We demonstrate that the proposed AE works robustly for all
interference levels. In particular, the performance improvement compared to conventional AE [7] is more notable for the strong and very strong interference scenarios.

\section{System Model}
\subsection{System description}

The proposed ADL algorithm based AE system for a wireless communication interference channel with $m$-user is shown in Fig. 1. It has three main blocks: transmitter, channel, and receiver. Compared to a conventional communication system with a number of blocks, this proposed diagram recast the block diagram as an end-to-end optimization task and represent the system as a simplified AE system by using a DL based neural network (NN) layer. For basics of DNN, an introduction is given in \cite{Goodfellow2016}. The NN layer stacks one on top of another. In general, the NN layers considered in this work transform an input data $l_{\text{in}}$ into an output $l_{\text{out}}$ as follows:
\begin{equation}
\label{eq:DNN transform}
l_{\text{out}}=f(\textbf{w}l_{\text{in}}+\textbf{b})
\end{equation}
where \textbf{w} and \textbf{b} are weights and trainable parameters and $f(.)$ is a non-linear function \cite{Goodfellow2016}. The weights of the whole layers are optimized jointly. Let $\textbf{\textit{s}}$ as the input, and the training set contains all the possible values of $\textbf{\textit{s}}$. In an AE during the training, the targets values are equal to the inputs $ie$ $\hat{s}_i=s_i$, where $s_i$ is a realization of $\textbf{\textit{s}}$. The network is trained to optimize the reconstruction error, which is given:

\begin{equation}
\label{eq:DNN reconstruction}
L(s,\hat{s})=-\text{log}p(s|\hat{s})
\end{equation}
The reconstruction error here is the cross entropy loss, which is given \cite{Goodfellow2016}:

\begin{equation}
\label{eq:cross entropy loss}
L(s,\hat{s})=-\sum_{k}(s(k)\text{log}\hat{s}(k)+(1-s(k))\text{log}(1-\hat{s}(k)))
\end{equation}
where $\hat{s}(k)=P(s(k)=1|\hat{s})$. $s(k)$ stands for bit $k$ of $s$ and $\hat{s}(k)$ stands for bit $k$ of $\hat{s}$. The training of the network is performed by solving the following optimization problem:

\begin{equation}
\label{eq:Optimization solver}
\operatorname*{arg\: min}_{P}\E_{s,N,\theta}[L(s,\hat{s})]
\end{equation}
where $P$ is denote the set of trainable parameters. $N$ and $\theta$ are generated noise and phase by the channel layer each time it is used. 

For the transmitter side, the transmitted messages $\textbf{\textit{s}}$ is reconstructed, and $s_i\in M=\{1,2, \ldots,M\}$, where $M=2^{k}$ is the dimension of $M$ with $k$ being the number of bits per message. The message is passed to the transmitter. The transmitter applies a transformation by a DNN layer $f:M \rightarrow$ $\mathbb{R}^{2n}$ to the message $s_i$ to generate the transmitted signal $x=f(s_i)\in\mathbb{R}^{2n}$. Note that the output of the transmitter is an $n$-dimensional complex vector which is transformed to a 2$n$ real vector. We use 'one-hot vector' with size of $M$ to reconstruct $s_i$ for DNN layer. Following the similar definition in [7], the transmitter is constrained by an average power: $\E{[|x_i^{2}|]}\leq 0.5\: \forall\: i$. In this work, we use a $m$-user interference channel with AWGN.

\begin{table}
\begin{center}
\caption{The structure of the MLP AE}
\label{tab:structure of AE}
\begin{tabular}{p{2.5cm}|p{3cm}|p{2.5cm}}
\hline\hline
Block name  & Layer name & Output Dim \\ 
\hline
            & input: & $M$ \\ 
Block name  & Dense+eLu & $M$ \\ 
            & Dense+Linear & $2n$ \\ 
            & nomalization & $2n$ \\ 
\hline
Channel     & Noise & $2n$ \\ 
\hline
Decoder     & Dense+ReLU & $M$ \\ 
            & Dense+Softmax & $M$ \\ 
\hline\hline
Name  & $[\sigma(u)]_i$ & range \\ 
\hline
ReLU     & max(0,\:$U_i$) & [0,\:$\infty$) \\ 
Tanh     & tanh($U_i$) & (-1,\: 1) \\ 
Softmax  & $\frac{e^{u_i}}{\sum_je(u_j)}$ & (0,\: 1) \\ 
\hline\hline
\end{tabular}
\end{center}
\end{table}

\subsection{Model of interference channel with $m$-user}

In Fig.1, a $m$-user Gaussian AWGN interference channel is illustrated within the dashed-line rectangle block. The interference channel has $m$ transmitter-receiver pairs that simultaneously communicate in blocks of size $m$. Each transmitter communicates to its own receiver a message $s \in m =\{1, 2, \dots, m \}$. Let $x^{n}$ and $y^{n}$ denote the input and output signal of the $n\:th$  user, respectively. $N^n \sim CN(0, 1)$ is independent and identically distributed Gaussian noise that impairs receiver $n$. Each $x_n$ has an associated average power constraint $P^n$ so that $\frac{1}{m}\sum_{n=1}^{m} |x^{n}_{m}|^{2}\leq P^n$. Receiver $n$ observes $\hat{y}^n$ and estimates the transmitted message $\hat{x}^n$. The average probability of error for user $n$ is $\epsilon^{n}_{m}=\E[P(\hat{s}^n \neq s^n)]$, where expectation is over the random choice of message.  The channel output at each receiver is a noisy linear combination of its desired signal and the sum of the interfering terms, of the form \cite{Jafar2010}:

\begin{equation}
\label{eq:Channel}
y^{n}=x^{n}+\sqrt{\frac{\text{INR}}{\text{SNR}}}\sum_{j=1,j\neq n}^{m} x^j+N^n,\forall j, n=1,2,\dots, m
\end{equation}
where $y^n$ and $N^n$ are the channel output and AWGN respectively, at the $n\:th$ receiver and the $x^n$ is the channel input symbol at the $n\:th$ transmitter. All symbols are real and the channel coefficients are fixed. 
The AWGN is normalized to have zero mean and unit variance and the input power constraint is given by \cite{Jafar2010}:

\begin{equation}
\label{eq:AWGN input}
E[(x^n)^2]\: \leq \:\text{SNR}, \:\: \forall \:n \in \:m.
\end{equation}
The INR is defined through the parameter $\alpha$ \cite{Jafar2010}:

\begin{equation}
\label{eq:INR}
\frac{\text{log}(\text{INR})}{\text{log}(\text{SNR})}=\alpha  \:\:\rightarrow \:\:\text{INR}=\text{SNR}^{\alpha}
\end{equation}

Note that the definition of INR ignores the fact that there are $m$-1 interferers observed at each receiver. This is for two reasons. First, this definition parallels that of the two-user case [20], which will make it easier to compare the two rate regions. Second, the receivers will often be able to treat the interference as stemming from a single effective transmitter, via interference alignment. This is not the case when the receiver treats the interference as noise. In this work, the introduced parameter $\alpha>0$ defined by $\text{INR}=\text{SNR}^{\alpha}$; this coupling parameter $\alpha$ is used to specify the corresponding linear deterministic model in [21]. 

In this work, we address the interference scenarios including noisy, weak, moderate, strong, and very strong interferences. The definition of the classification for the interference is proposed in \cite{Jafar2010}. The degrees-of-freedom (GDoF) of the symmetric $m$-user interference channel is identical to that of the multiple-user channel, except for a singularity at $\alpha$ = 1, as follows:

\begin{equation}
\label{interference classification}
d(\alpha)=
\begin{cases}
\; 1-\alpha, \;\:  0 \leq \alpha < \frac{1}{2} \: (\text{noisy}) \\
\; \alpha, \: \qquad   \frac{1}{2} \leq \alpha < \frac{2}{3} \: (\text{weak}) \\
\; 1-\frac{\alpha}{2}, \,\:  \frac{2}{3} \leq \alpha < 1 \: (\text{moderate}) \\
\; \frac{1}{K}, \qquad \quad \alpha=1 \\
\; \frac{\alpha}{2},\: \qquad   1 < \alpha < 2 \: (\text{strong}) \\
\; 1, \:\: \qquad   \alpha \geq 2  \,\qquad  (\text{very strong}) \\
\end{cases}
\end{equation}

\begin{algorithm}[t]

\label{alg:DRL algorithm}
   \caption{ADL algorithm to predict the interference}
    \SetKwInOut{Input}{Input}
    \SetKwInOut{Output}{Output}
    
    \Input{$\bullet$ AE model and specifications: $n$, $k$, batch size, epochs number, optimizer, learning rate, etc \\ 
    $\bullet$ the training data set $l_{\text{in}}$\\
    $\bullet$ the variance of channel noise $\sigma^2$ \\}
    \Output{$\bullet$ •	the estimated interference parameter $\alpha$\\}

Initialize:\\
       Set AE model parameters (e.g., $n\leftarrow$4,\:$k\leftarrow$4,\:$M\leftarrow$4)\\ 
       \textbf{for} $i$ in range (training data samples)  \textbf{do}\\
       \quad Set $x=f(s_i )\in$ $\mathbb{R}^{2n}$,  $s_i \in \{1,2\dots M\}$, encoding\\
       \quad Create and Set $\hat{y}(n)$ for receiver layer\\
       \quad \textbf{for} $i$ in range (numble of guessing $\alpha$)  \textbf{do}\\
       \qquad DNN layer to training data set (settings in Table I)\\
       \qquad Recovery pilot signal $\hat{s}_i$ according to a guessing $\alpha$\\
       \qquad Calculate \textit{reward} $\hat{R}_i$ according to Eqs. (5) and (6)\\
       Set confidence interval of $\hat{R}_i$ and predict $\alpha$\\
       Update DNN layer with $\alpha$ according to Eqs. (7) to (10) 
\end{algorithm}

\subsection{ADL algirthm at receiver blocks}

As shown in Fig.1, at the receiver side, $y(n)$ is the received signal after propagating through an AWGN channel, which includes the original transmitted signal, the channel response, AWGN noise as well as the interference from other sources. Here, the received $n$-dimensional signal $y(n)$ noised by a channel represented as a conditional probability density function $p(y|x)$, and the DNNs receiver subsequently learns it with multiple dense layers. The last layer of the receiver is a Softmax activation layer that outputs an $M$-dimensional probability vector $\textbf{\textit{p}}$, in which the sum of its elements is equal to 1. The receiver first applies the transformation $f:\mathbb{R}^{2n} \rightarrow M$ to decode the signal, creating a signal $\hat{s}_i$   to recover the original transmitted signal $s_i$. 

To enable the comparability of the results implemented in different scenarios, we set $n$=4 and $k$=4 throughout this work. For other setups of the AE, to allow a benchmark for comparison, we use the similar AE structure and settings as in [7], which are based on a multi-layer perceptron (MLP) AE. The specifications are listed in Table I. We train the AE in an end-to-end manner using the Adam optimizer, on the set of all possible messages $s_i \in M$, using the cross-entropy loss function. ReLU and Softmax are used in DNNs layer.

As shown in Fig.1, we design and propose an adaptive learning processing, integrating with the DNNs based receiver block, named \textit{ADL algorithm}, to estimate the interference coupling parameter $\alpha$. With the Predicted $\alpha$, we obtain an updated channel function, according to Eqs. (5) to (7). Then the DNN layer is updated with this knowledge by substituting $\alpha$ into Eq. (5). This process includes two stages. Firstly, we utilize multiple group of pilot signals for online DNN training to predict the real-time $\alpha$. Then with the knowledge of the channel, we update the interference channel function, decode signals with DNN layers. 

It assumes that the signals consist of two parts. The first part is pilot signal, as the training data set. The second part is the transmitted signal, which has the same structure as it’s in a DL based OFDM system [17]. However, we utilize the pilot signals here for both estimating interference and the DNN training. We introduce and explain our proposed ADL algorithm in Algorithm I. At the initialization stage, we set the specifications of an ($n$=4,$k$=4) AE and load the input training data set. Then, the DNN layer encodes the data for propagating through an AWGN channel. The DNNs based receiver block first captures a group of signals, and then the reinforcement block starts to train the pilots simultaneously. By process of reward computation, the block normalizes the reward regarding different guessing values of $\alpha$. Then we determines the optimum $\alpha$ range with regarding the a predefined confidence interval. Based on the plot of the reward according to the guessing values of $\alpha$. We compute the mean, as the predicted $\alpha$. Next, the estimated $\alpha$ is substituted back into the DNNs block for the decoding process with an updated DNNs layer. For this prediction process based on the reward performance, we will give more details in the Section of Numerical Evaluation.  In this work, the normalized reward is defined as follows:  

\begin{equation}
\label{eq:normalized reward}
\hat{R}_i= \frac {R_i}{||R_i||} 
\end{equation}
where
\begin{equation}
\label{eq:reward}
R_i= \frac {1}{\overline{BER}|_{Pilot\leftarrow(1,\dots,i)}} 
\end{equation}
$R_i$ is defined as the reciprocal of the mean bit error rate (BER) value for $i$ pilots signals. 

\begin{figure}
\begin{center}
\includegraphics[width=.75\textwidth]{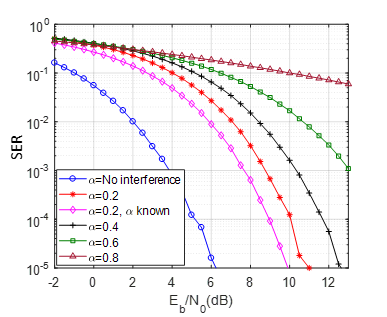}
\end{center}
\caption{SER versus SNR performance of an AE (4, 4): no interference, interference $\alpha$=(0.2 0.8) with blind training, $\alpha$=0.2 with knowledge of $\alpha$. 
\label{fig:BER_no_interference}}
\end{figure}

\begin{figure}
\begin{center}
\includegraphics[width=.75\textwidth]{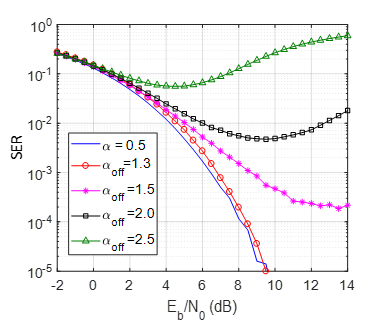}
\end{center}
\caption{SER versus SNR performance of an AE (4, 4): weak interference $\alpha=0.5$ (at training) with offset up to $\alpha_{\text{off}}=2.5$ (received).
\label{fig:weak_interference}}
\end{figure}

\section{Numerical Results and Discussion}
In this section, numerical simulation is carried out under the environment of Python 3.0, with the libraries of PyTorch, TorchNet and TQDM. Training was done at a fixed value of $E_b/N_0$ = 7 dB using Adam [22] with a learning rate of 0.001. Activation functions rectified linear units (ReLU) [23] and Softmax are used  in our DNNs layer. The details are listed in Table I. Detailed explanation of these can be found in [24]. The pilot symbol ratio we used in our simulation is 0.01. The group number of the bit streams is 30, which is used for jointly training and estimating the interference $\alpha$. 

\subsection{Comparsion with and without interference}

A DL based AE with different settings of ($n$, $k$) ($n$ is the number of channel use, and $k$ is the bits of the signal) are studied and evaluated in [7]. It compares the performance between the $M$-QAM modulation and AE with similar settings. It demonstrates that the AE (4, 4) and (4, 8) outperforms the 4-QAM and 16-QAM. To enable a benchmark for comparison, we choose the setting (4, 4) throughout all scenarios. However, we evaluate the performance according to our proposed interference model, as shown in Eqs. 5-8. We verify our algorithm through an example of a two-user interference channel case. For other multi-user case, the methodology is similar, and the enhancement is more significant.

In the proposed DL based system, the AE reconstruct and compressed the data with 'one hot vector' format for the NN layer. For a fair comparison to a conventional system with other modulation schemes, we study the symbol error rate (SER) for evaluating the system performance. We first simulate an AE (4, 4) system in an ideal channel without taking any interference, as a reference point. The plot is illustrated in Fig.2. It shows that without the interference, the system works well, even under a low SNR. Then we evaluate a blind training with interference. The blind training is defined as that the system does not have any knowledge that it is an interference channel. Therefore the system trains a model without interference $\alpha$. However, the true received signal has a certain value of $\alpha$. We evaluate the system from $\alpha=0.2$ to $\alpha=0.8$, in Fig.2. The results show that with a blind training, the AE has some robustness even without any knowledge of the channel. However, when $\alpha$ increases beyond 0.6, then the AE doesn’t work well.  We also plot the case with the knowledge of $\alpha$ for comparison. When $\alpha=0.2$, we could achieve SER $\sim 10^{-3}$ at $E_b/N_0 =\: \sim7$ dB, and this is assuming that we know the exact $\alpha$ for training. The comparison in Fig. 2 indicates that it is possible to overcome the interference effect if we have an efficient approach to predict the interference parameter $\alpha$. 

\subsection{Robustness of an AE for different interference strengths}

We demonstrate that the AE approach has some robustness when it applies in an interference channel. However, we also want to characterize the robustness for difference interference strengths. It assumes that the system knows the interference channel generalized formula (Eq.5) and it applies a DL training for the decoder. We train the model with a predetermined $\alpha$. However, we assume that $\alpha$ may change dynamically in a real time scenario and we want to evaluate how robust of the decoder when $\alpha$ has some offset, denote as $\alpha_\text{off}$. 

 Following the definition in Eqs.5 to 8 , we simulate for weak ($\alpha=0.5$) and very strong interference ($\alpha=2$) respectively.  Results are plotted in Figs. 3 and 4. It shows that the AE approach is quite robust for a weak interference. The system works even under a very large offset: 3 times of the training $\alpha$. However, the situation is slightly different for very strong interference, where $\alpha=2$. The result in Fig. 4 indicates that the system is quite sensitive to the offset under a very strong interference channel. For this scenario, it does require a technique to deal with the interference. To address this, we apply the proposed ADL algorithm and the performance evaluation is given in the next section. 
 
 \begin{figure}
\begin{center}
\includegraphics[width=.75\textwidth]{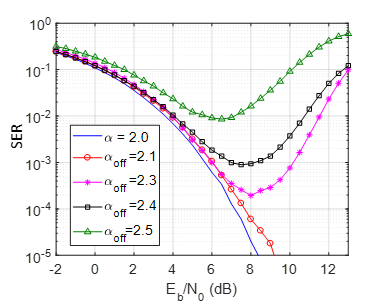}
\end{center}
\caption{SER versus SNR performance of an AE (4, 4): very strong interference $\alpha=2$ (at training) with offset up to $\alpha=2.5$ (received). 
\label{fig:very_strong_interference}}
\end{figure}

\begin{figure}
\begin{center}
\includegraphics[width=.75\textwidth]{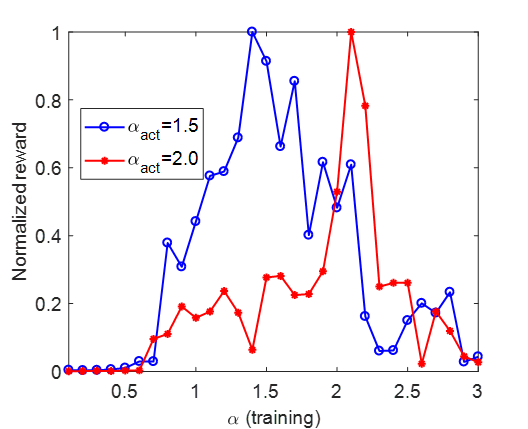}
\end{center}
\caption{Normalized reward versus predicted $\alpha$: strong interference $\alpha=1.5$ and very strong interference $\alpha=2$. 
\label{fig:normalized_reward}}
\end{figure}

\begin{figure}
\begin{center}
\includegraphics[width=.75\textwidth]{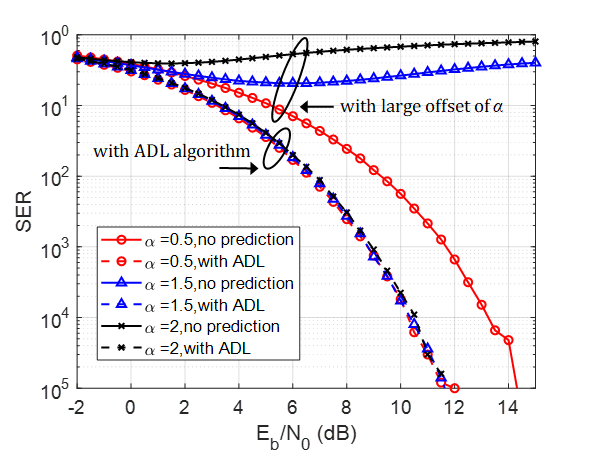}
\end{center}
\caption{SER versus SNR: comparison for strong and very strong interference channel, with and without the proposed ADL algorithm. 
\label{fig:comparison_with_without_DRL_algorithm}}
\end{figure}

 \subsection{Evaluation with the proposed learning algorithm}
 
 Recall the proposed ADL algorithm in section II. We evaluate the ADL algorithm to estimate $\alpha$ in different interference strengths. We also carried more groups of study as in section B, and we found that for strong ($\alpha=1.5$) and very strong ($\alpha=2$) interference, the offset of $\alpha$ becomes more critical. Therefore, we address this and implement our algorithm for these cases. With the same setting in Figs. 3 and 4, we plot the normalized reward versus a predicted $\alpha$ (different values at training), in Fig. 5. for $\alpha=1.5$ and $\alpha=2$, respectively.  We can see that the peak value of the normalized reward appears around 1.5 (actual value), and it reduces gradually to both sides of the actual value. By contrast, for the very strong interference, where $\alpha=2$, we can also found out the peak value of the normalized reward appears around the real value of $\alpha$. However, it decreases rapidly towards both sides of the actual value, which agree with the achievement that it’s more sensitive to the offset. As the fluctuation is quite large in Fig. 5, here we define 40\% offset as the confidence interval of the reward, to estimate $\alpha$. We use the mean $\alpha$ for evaluating the performance, as we introduced in Section II. Furthermore, the reward is computed according to the instant SNR condition. For this simulation, we use $E_b/N_0 = 7$ dB as an example. To evaluate the performance with and without applying the proposed ADL algorithm, we plot the SER performance for weak, strong and very strong interference channels for comparison, as shown in Fig. 6. In this simulation, we take a large interference effect as an example, $\alpha_{\text{off}}=2\alpha$, to demonstrate the improvement achieved by our algorithm. Two groups of data are highlighted in Fig. 6. We can see that the SER significantly degrades due to the large offset of $\alpha$. In particular, for the strong and very strong interference cases, the system does not work without the knowledge of $\alpha$. However, with applying the ADL algorithm, the result shows that with an efficient interference prediction, the ADL algorithm based AE is capable of robust performance over the entire range of interference levels, even for the worst case in a very strong interference channel. 
 
 \section{Conclusion}

An ADL algorithm based AE is proposed for interference channel with unknown interference. With the proposed online learning, interference can be estimated and predicted, which is then subsequently used for decoding of the signals using DNN. The proposed algorithm is shown to significantly enhance the robustness of the interference channel, and provides an AE system that is adaptable to real-time interference, for the entire range of interference levels. The enhancement is more notable for strong and very strong interference scenarios, compared to performance of conventional AE with offline learning.  

We believe that our proposed approach is an important step towards enabling AE for 5G and beyond communication systems with dynamic and heterogeneous interference. Our future work aims at improving computational efficiency of our online learning scheme, and the implementation on real-life platforms. 

%
%
%
%

\end{document}